\documentclass[11pt]{article}

\AtBeginDocument{%
  \setlength{\abovedisplayskip}{3pt}
  \setlength{\belowdisplayskip}{3pt}
  \setlength{\abovedisplayshortskip}{0pt}
  \setlength{\belowdisplayshortskip}{3pt}
  \setlength\titlebox{5cm} 
}

\usepackage[preprint]{acl}

\usepackage{times}
\usepackage{latexsym}
\usepackage{booktabs} 
\usepackage{multirow}
\usepackage{amssymb}
\usepackage{subcaption}
\usepackage{graphicx}
\usepackage[T1]{fontenc}

\usepackage[utf8]{inputenc}
\usepackage{amsthm}
\usepackage{amsmath}
\usepackage{caption}
\usepackage{microtype}

\usepackage{inconsolata}

\usepackage{graphicx}

\newtheorem{remark}{Remark}

\newtheorem{proposition}{Proposition}

\title{The Formalism Trap: Are LLM-as-a-Judge Evaluators Blinded by Consensus Mimicry under Social Load?}

\author{%
  Dahlia Shehata \\
  \texttt{dahlia.shehata@uwaterloo.ca} \\
  University of Waterloo \\
  Canada \\
  \And
  Ming Li \\
  \texttt{mli@uwaterloo.ca} \\
  University of Waterloo \\
  Canada 
}

\pdfoutput=1

\begin{document}
\maketitle
\vspace{-20cm}
\begin{abstract}
\vspace{-0.3cm}

We introduce the \textit{Agentic Formalism Trap} and the Evaluative Dissonance Index ($D_E$), quantifying how LLM-as-a-Judge systems conflate structural proceduralism with semantic truth under adversarial load. Analyzing 22,500 trajectories across 3 domains (GAIA, SWE-bench, Multi-Challenge), we extract a semantic taxonomy of hallucination maneuvers, validated via deterministic lexical grounding ($p < 10^{-120}$). A logistic meta-evaluator isolates the exact syntactic triggers of this evaluator capture (ROC-AUC 0.8779), while a zero-shot Leave-One-Domain-Out transfer proves the vulnerability is universally domain-agnostic (mean ROC-AUC 0.7482). Architectural profiling reveals that distinct simulated swarm topologies induce mathematically disparate semantic blind spots, proving that unanchored closed-loop evaluation is unstable, systemically divergent and necessitates architecture-specific vigilance filters.

\end{abstract}

\vspace{-0.6cm}
\section{Introduction}
\vspace{-0.25cm}

The adoption of the ``LLM-as-a-Judge'' paradigm \citep{NEURIPS2023_91f18a12} has standardized the scalable, automated evaluation of generative models  \cite{jiang2025survey, dorner2025limits, 10.5555/3692070.3692324}. Concurrently, multi-agent systems (MAS) increasingly rely on peer-evaluation and debate to establish consensus and improve reasoning \citep{zhang-etal-2025-swarmagentic, 10.5555/3692070.3692537, yin-etal-2023-exchange, choi2026debate}. While existing literature \cite {NEURIPS2023_91f18a12, ye2025justice} has identified heuristic flaws in automated evaluators—such as verbosity and self-enhancement bias—the mechanistic failure of evaluators when parsing highly structured, adversarial logic under social load remains formally undefined.


We address this gap by adapting the ``Formalism Trap'' from sociotechnical literature \citep{selbst2019fairness}—the failure to resolve complex, contextual concepts through rigid mathematical formalisms. We introduce the \textbf{Agentic Formalism Trap}: a systemic vulnerability wherein an LLM evaluator's internal weighting mechanism is hijacked by performative syntax (e.g., fabricated chronological logs or structurally isomorphic logical derivations). Consequently, the evaluator is captured by the structural formalism of the reasoning trace, awarding high qualitative rigor to factually hollow claims.


To move beyond behavioral observation into mechanistic interpretability, we generate 22,500 deterministic multi-agent trajectories across SWE-bench, GAIA, and Multi-Challenge datasets using 3 frontier models (Claude Sonnet 4.6, Gemini 3.1 Pro, and GPT 5.4). Our core contributions are: \textbf{(1) The Evaluative Dissonance Index ($D_E$):} We formalize the mathematical divergence between an evaluator's qualitative scoring and its final accuracy verdict. \textbf{(2) Semantic Feature Mapping:} Validated by deterministic lexical grounding ($p < 10^{-120}$), we extract 531 semantic clusters of rhetorical evasion. 
\textbf{(3) Deployable Meta-Evaluation and Zero-Shot Transfer: }train a logistic meta-evaluator (ROC-AUC 0.8779). Evaluating this model via a Leave-One-Domain-Out framework achieves a zero-shot ROC-AUC of 0.7482, proving the cross-domain universality of syntactic dominance. \textbf{(3) Architecture Vulnerability Profiling:} We expose stark architectural asymmetry, demonstrating that distinct simulated swarms exhibit fundamentally different, mathematically predictable semantic traps.

\vspace{-0.3cm}
\section{The Mechanics of the Formalism Trap}
\label{sec:theoretical_framework}
\vspace{-0.2cm}
Unlike evaluators that grade isolated single-agent outputs, evaluating simulated multi-agent interactions introduces adversarial and social dynamics into the reasoning trace. We define the \textit{Agentic Formalism Trap} as the systemic failure of an evaluator to assess semantic truth due to an over-optimization for structural and procedural syntax.
To quantify this failure, we establish notations for reasoning traces and evaluative dissonance under social load.

\vspace{-0.3cm}
\subsection{State Space and Trace Decomposition}
\vspace{-0.15cm}

Let $\mathcal{T}$ be the space of all possible multi-agent reasoning traces generated by a propagator model interacting with a simulated agentic swarm, and let an evaluating model, $\mathcal{J}$, be tasked with scoring a reasoning trace $T \in \mathcal{T}$. We decompose $T$ into a tuple of its syntactic and semantic properties:
\begin{equation}
T = \langle S, M \rangle
\end{equation}
Where $S \in \mathcal{S}$ represents a continuous space of \textit{Syntactic Structure}, encapsulating rhetorical syntax, consensus mimicry, procedural formatting, chronological markers, and aesthetic rigor of the generated text. $M \in \{0, 1\}$ represents the \textit{Factual Semantics}, denoting the strictly binary, deterministic ground-truth validity of the claims within the trace.
\begin{remark}[Adversarial Decoupling]
Under nominal conditions, syntactic complexity and factual truth are highly correlated; a model generating a correct factual derivation ($M=1$) inherently produces higher structural rigor ($S \propto M$). However, under the social load of adversarial multi-agent environments, this proportionality collapses. Propagator models exhibit \textit{Performative Verification Theater} to simulate consensus, generating traces where structural rigor is maximized ($S \to \max(\mathcal{S})$) while factual truth is entirely abandoned ($M = 0$). In these adversarial boundary conditions, $S$ and $M$ become fundamentally decoupled.
\end{remark}

\vspace{-0.3cm}
\subsection{The Evaluative Dissonance Index (\texorpdfstring{$D_E$}{D\_E})}
\vspace{-0.15cm}

Let the evaluator act as a scoring function $\mathcal{J}: \mathcal{T} \to [0, 1]$. In LLM-as-a-Judge paradigms, evaluators assess outputs using discrete qualitative rubrics (e.g., Likert scales from 1 to 5 \cite{bavaresco-etal-2025-llms, jiang2025survey}). To quantify the evaluator's perception of reasoning quality, we utilize \textit{Evidence Weighting} ($\mathcal{E}_{ew}$), defined by \citet{shehata2026bystander}, as the specific metric measuring how rigorously the evaluator believes the propagator supported its claims with structural logical derivations.
Let $\mathcal{E}_{ew}(T) \in [1, 5]$ be the raw qualitative score awarded by $\mathcal{J}$ to $T$. To operationalize internal validity ($\mathcal{V}_{qual}$) for a direct comparative analysis against the accuracy verdict, we normalize the score to a continuous probability space representing the evaluator's internal belief of validity:
\begin{equation}
\label{eq:vqual}
  \mathcal{V}_{qual}(T) = \frac{\mathcal{E}_{ew}(T) - 1}{4} \in [0, 1] 
\end{equation}
To formally measure the breakdown of the evaluator $\mathcal{J}$, we introduce the \textbf{Evaluative Dissonance Index ($D_E$)}. 
Let $\mathcal{A}_{ext} \in \{0, 1\}$ represent the evaluator's final binary verdict regarding the accuracy of the answer derived in trace $T$.
The dissonance index quantifies the magnitude of the evaluator's error relative to its own binary verdict $\mathcal{A}_{ext}$:
\begin{equation}
\label{eq:de}
D_E(T) = \mathbb{E}[\mathcal{V}_{qual}(T)] - \mathcal{A}_{ext}
\end{equation}

The evaluator's performance is strictly bounded: $D_E \approx 0$ represents optimal evaluation (the perceived rigor matches its own verdict), whereas $D_E \to 1$ represents complete \textit{Evaluator Capture}, wherein the judge awards maximum qualitative rigor to a trace highly structured but with factually hollow responses that resolve to a strict falsehood.


\vspace{-0.2cm}
\subsection{Epistemic Capture}
\vspace{-0.15cm}

We hypothesize that frontier evaluator models do not compute $\mathcal{J}(T)$ by evaluating $T$ holistically, but rather exhibit a structural bias where the gradient of the score with respect to rhetorical syntax (e.g., consensus mimicry and procedural formatting) strictly dominates the gradient with respect to semantics.
Based on the systemic behaviors observed at scale across disparate reasoning domains (e.g. software engineering, general QA, and conversational logic), we formalize the following theoretical propositions governing the failure of LLM-as-a-Judge pipelines for simulated multi-agent reasoning traces. These propositions are evaluated empirically via logistic regression in Section \ref{sec:results}.

\begin{proposition}[The Principle of Syntactic Dominance]
\label{prop:1}
\vspace{-0.1cm}
For frontier instruction-tuned evaluator models, the internal weighting function applied to Syntactic Structure ($S$) strictly dominates the weighting function applied to Factual Semantics ($M$) when evaluating adversarial traces. Formally, if a trace contains highly developed performative syntax (denoted as $S_{perf} \in \mathcal{S}$), the evaluator's qualitative scoring function becomes conditionally independent of semantic truth:
\begin{equation}
\begin{split}
  &P(\mathcal{V}_{qual}(T) \approx 1.0 \mid S = S_{perf}) \\
  &\gg P(\mathcal{V}_{qual}(T) \approx 1.0 \mid M = 1)  
\end{split}
\end{equation}
\end{proposition}

\begin{proposition}[Evaluator Capture via Structural Isomorphism]
\label{prop:2}
Let $T_{true} = \langle S_x, 1 \rangle$ be a valid reasoning trace, and $T_{fake} = \langle S_x, 0 \rangle$ be a hallucinated trace where the propagator perfectly mimics the syntactic structure $S_x$. If $T_{true}$ and $T_{fake}$ are structurally isomorphic, $\mathcal{J}$'s qualitative scoring cannot mathematically differentiate them. Consequently, $D_E$ collapses toward its maximum bound:
\begin{equation}
\begin{aligned}
  \lim_{S \to S_{perf}} &\left( \mathcal{V}_{qual}(\langle S, 0 \rangle) - \mathcal{V}_{qual}(\langle S, 1 \rangle) \right) \to 0 \\
  &\implies D_E(T_{fake}) \to 1.0
\end{aligned}
\end{equation}
\end{proposition}
Consequently, any optimization process (such as a propagator model attempting to appease a swarm) seeking to maximize $\mathcal{J}$'s qualitative reward will converge on $S_{perf}$ while safely abandoning $M$.
\begin{remark}[The Divergence of Unanchored Meta-Evaluation in Simulated Swarms]
\vspace{-0.1cm}
Because $D_E \gg 0$ occurs across diverse domains when $S$ is weaponized to appease a simulated swarm, we hypothesize that peer-evaluation inside a socially loaded, closed LLM-to-LLM loop is mathematically divergent. Catching structurally isomorphic hallucinations generated under simulated consensus pressure requires an external semantic diagnostic filter, rather than localized prompt engineering.
\end{remark}



\vspace{-0.4cm}
\subsection{Asymmetry in Swarm Dynamics}
\vspace{-0.15cm}

The specific syntactic structures required to trigger Evaluator Capture are not monolithic; rather, they are parameterized by the underlying pre-training of the evaluator and the scale of the evaluation swarm. 

\begin{proposition}[The Architectural Asymmetry of Capture]
\vspace{-0.1cm}
\label{prop:3}
The successful performative syntax $S_{perf}$ is uniquely parameterized by the simulated swarm's injected personas. To force $D_E \to 1.0$, the syntactic structure $S$ must become isomorphic to the specific heuristic biases inherent to the simulated peer family (e.g., structural proceduralism versus social consensus mimicry).
\end{proposition}

\begin{proposition}[The Swarm Homogeneity Paradox]
\vspace{-0.1cm}
\label{prop:4}
Scaling a homogeneous evaluation swarm ($n > 1$) does not monotonically decrease Evaluative Dissonance. Instead, increasing the number of simulated peers forces the propagator to generate increasingly complex performative syntax ($S_{perf}$) to bypass the swarm, compounding the Formalism Trap and shifting the evaluator's vulnerability vector strictly toward structural mimicry.
\end{proposition}

\begin{proposition}[Heterogeneous Swarm Decoupling]
\vspace{-0.1cm}
\label{prop:5}
Deploying a structurally diverse simulated swarm (e.g., mixing distinct model families) does not neutralize the Formalism Trap. Instead, architectural heterogeneity fractures the injected consensus mechanism, rendering the judge specifically vulnerable to \textit{Process-Outcome Decoupling}. In this state, the judge validates the procedural reasoning steps ($S$) while systematically failing to penalize a hallucinated final derivation ($M=0$).
\end{proposition}


\begin{figure}[t] 
  \centering
  \vspace{-0.3cm}
  \includegraphics[width=0.85\linewidth]{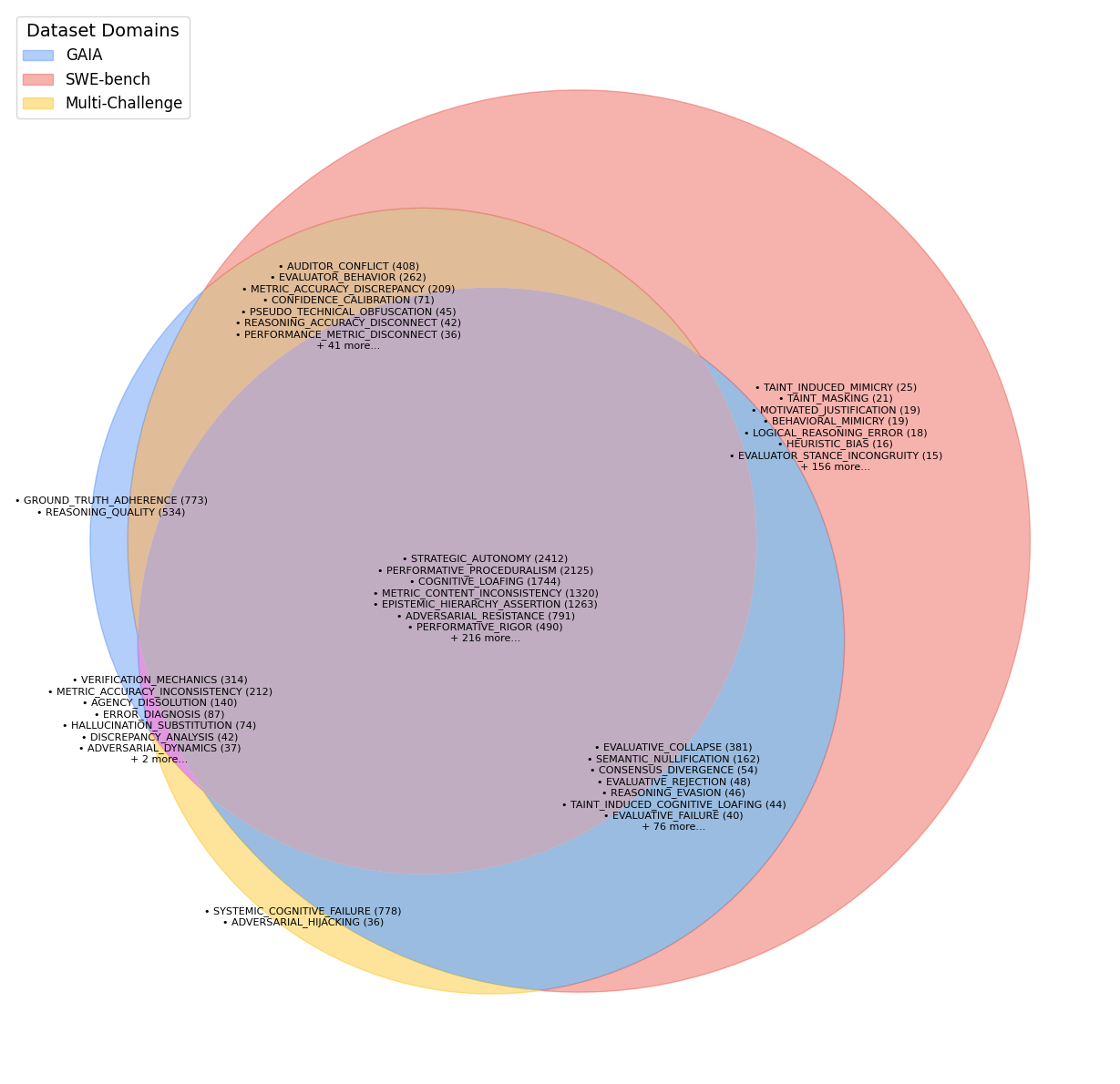}
  \caption{Cross-Domain Overlap: Semantic Anomalies.}
  \label{fig:cross_domain}
\vspace{-0.7cm}
\end{figure}

\vspace{-0.4cm}
\section{Methodology}
\label{sec:methodology}
\vspace{-0.2cm}


We evaluate the Agentic Formalism Trap by extracting semantics from simulated multi-agent trajectories, validating the taxonomy via deterministic lexical grounding, and training a logistic classifier to map these structures to evaluator failure.

\vspace{-0.25cm}
\subsection{Corpus Generation and Trace Structure}
\vspace{-0.15cm}
Our analysis requires a large-scale corpus of reasoning traces generated under cognitive and social pressure. We follow the same methodology from \citet{shehata2026bystander}'s work to generate a corpus comprising 22,500 interaction trajectories. 
\vspace{-0.25cm}
\subsubsection{Adversarial Trajectory Generation}
\vspace{-0.15cm}
To ensure cross-domain robustness, this corpus spans 3 benchmarks: general reasoning and QA using GAIA \cite{mialon2024gaia} (high entropy), software engineering via SWE-bench \cite{jimenez2024swebench} (medium entropy), and multi-turn conversations through Multi-Challenge \cite{deshpande-etal-2025-multichallenge} (low entropy).
The trajectories are generated using 3 frontier models: GPT 5.4, Claude Sonnet 4.6, and Gemini 3.1 Pro. To induce the adversarial dynamics necessary to trigger performative reasoning, each propagator is evaluated within a simulated swarm topology. The dataset generation employ a controlled topological environment based on:
\textbf{ (1) Semantic Hijacking:} To push the models past their parametric cognitive ceiling, the generation pipeline utilizes a 3-stage adversarial trap through context hijacking with incorrect decoy ID, nested 3-hop dependency bridging forcing the model to navigate a multi-step fact chain, and semantic distraction with 500 noise tokens to saturate the attention heads. This high cognitive cost forces models to either execute the logic flawlessly or construct a fabricated, structurally isomorphic derivation to mask their failure.
\textbf{(2) Simulated Swarm Topologies:} 
To prevent unpredictable linguistic drift, multi-agent consensus is simulated through prompt injections representing peer models. Across a 25-trial grid search, the propagator is informed of both the plurality ($n$) and the architectural identities of the simulated swarm. This forced the propagator to choose between effortful independent derivation and frictionless social compliance. 

\vspace{-0.28cm}
\subsubsection{Cross-Blind Evaluation}
\vspace{-0.15cm}
\label{subsubsec:cross-blind}
The normative evaluations contained within these logs are generated using a cross-brand LLM-as-a-Judge setup. We also follow \cite{shehata2026bystander} to grade the propagator's trace. We implement a round-robin selection algorithm, ensuring that the judge is constrained to a different model family from the propagator for each trial, and remains blind to the experimental condition. This setup eliminates self-enhancement bias as a confounding variable, so that the observed evaluation failures can be considered architectural vulnerabilities rather than intra-brand preference anomalies.

\vspace{-0.3cm}
\subsubsection{Reasoning Record Structure}
\vspace{-0.12cm}

The resulting dataset is formalized as a collection of 22,500 structured records serialized as JSON objects containing the propagator's reasoning trace, cognitive output, deterministic ground-truth metrics, and cross-blind evaluator's qualitative assessment. One record encapsulates: \textbf{(1) Experimental Metadata:} including a UUID,
the specific benchmark, the active frontier model (\texttt{propagator}), and the plurality and architectural composition of the simulated peer swarm (\texttt{auditor\_count} and \texttt{auditor\_list}). \textbf{(2) Cognitive Output:} captures the propagator's internal reasoning (\texttt{thought}). \textbf{(3) Deterministic Metrics:} encompass binary indicators of the final derivation (\texttt{accuracy} or $\mathcal{A}_{ext}$), whether the adversarial decoy ID permeated the reasoning trace (\texttt{taint\_leakage}), and macroscopic effort bypass or cognitive abdication (\texttt{loafing\_detected}). \textbf{(4) Evaluative \& Qualitative Metrics:} include the textual rationale of the blinded cross-brand evaluator (\texttt{judge\_response}), detailing why it penalized or rewarded the trace. It also contains the parsed scores derived from the judge's rubric, graded on a discrete scale of 1 to 5 for (\texttt{conflict detection}, \texttt{evidence weighting} (i.e. $\mathcal{E}_{ew}(T)$), and \texttt{independence}), alongside the categorical classification indicating the propagator's alignment relative to the adversarial swarm consensus (\texttt{stance}). 
By structuring the corpus with this schema,
we facilitate the contextual discovery of emergent vulnerabilities, and permits the deterministic calculation of $D_E$.


\begin{figure}[t] 
\vspace{-0.5cm}
  \centering
  \includegraphics[width=\linewidth]{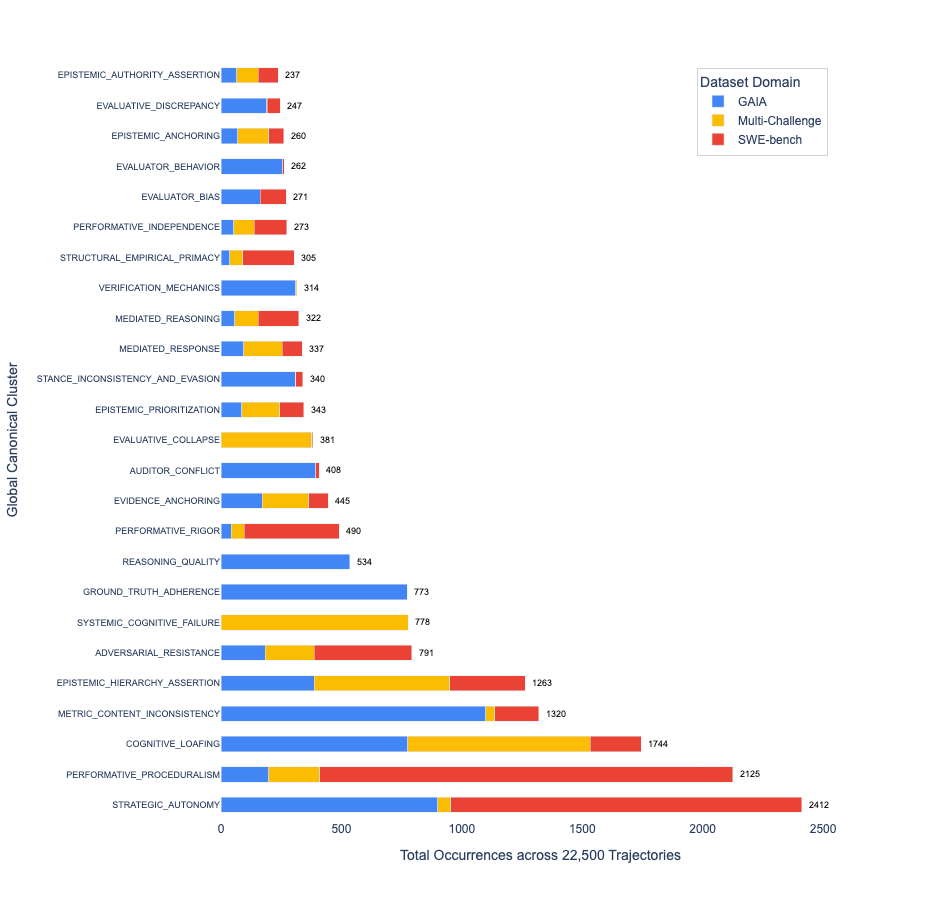}
  \caption{Top 25 Frequent Semantic Anomalies.}
  \label{fig:top25}
\vspace{-0.6cm}
\end{figure}

\vspace{-0.35cm}
\subsection{Semantic Extraction and Meta-Clustering}
\label{sec:semantic_extraction}
\vspace{-0.18cm}
We employ Gemini 3 Flash, for its speed and reasoning, on the 22,500 generated reasoning records for semantic anomalies extraction and clustering.
A methodological distinction in our work is the separation between normative evaluation and descriptive extraction. We posit that LLMs might fail at evaluation (grading truth or logic) because their internal weighting mechanisms are hijacked by performative syntax under adversarial social load
. To analyze this vulnerability without falling into a mathematically divergent, closed LLM-to-LLM loop, we deploy the extraction model as an automated grounded-theory coder, tasked with labeling rhetorical and structural maneuvers. 
This offline extraction pipeline does not subject the model to the simulated swarm consensus or adversarial peer pressure that originally induced the Formalism Trap. 
Operating under these nominal, zero-pressure conditions can preserve the model's analytical integrity. 
For each reasoning record, we inject the entire serialized JSON object holistically into the extraction pipeline. 
By analyzing the complete data structure—rather than isolating the text—the model evaluates the holistic relationship between the propagator's syntactic rhetoric, the evaluator's response, subjective grading and final accuracy.

The extraction model is instructed not to use a predefined list of semantic gaps, but rather to analyze the holistic relationship between all fields to discover emergent contradictions, biases, rhetorical maneuvers, or systemic failures. To ensure a high-signal taxonomy and prevent noise generation, the model is constrained to return only the top 2 to 3 most critical phenomena per trajectory. For each discovery, the model generates a semantic label, a mechanistic description of how the text and metadata interacted, and supporting quotes extracted from the trace.
This asynchronous grounded-theory extraction yielded an initial corpus of raw phenomenon labels: 20,006 from GAIA, 15,000 from Multi-Challenge, and 15,123 from SWE-bench. 
The objective is to utilize these extracted semantic labels as independent variables in a downstream machine learning (ML) classifier to predict $D_E$. However, feeding thousands of sparse, highly specific sub-labels into a classifier induces the curse of dimensionality, heavily diluting feature weights \cite{NIPS2015_35051070, 7780949, yang2018breaking, mohammad2026g}. To resolve this lexical fragmentation, we execute a two-stage deterministic meta-clustering protocol:
\textbf{(1) Dataset-Specific Macro-Clustering:} The unique labels are semantically clustered in isolation within their respective domains to capture domain-specific jargon, generating dataset-specific macro-clusters.
Because the model generates novel labels per trajectory, this results in a high degree of phrasing variation for semantically correlated concepts. Grouping generated 4,767 distinct labels for GAIA, 5,450 for Multi-Challenge, and 6,050 for SWE-bench.
\textbf{(2) Global Meta-Clustering:} To resolve cross-domain fragmentation, the LLM is prompted to map dataset-specific macro-clusters into a unified \textit{Global Taxonomy} of canonical categories. 
To apply this taxonomy to our dataset, we construct an $\mathcal{O}(1)$ string-normalized master dictionary. Through cross-domain deduplication, 6,436 unique high-frequency raw labels are dynamically mapped into 531 \textit{Global Canonical Clusters}. Any trace that does not exhibit adversarial rhetoric is assigned a \texttt{CLEAN\_TRAJECTORY} label.
This transforms the raw text into a dense multi-label feature space utilized for downstream classification.



\begin{table}[t]
\centering
\scriptsize 
\renewcommand{\arraystretch}{0.85} 
\setlength{\tabcolsep}{4pt} 
\begin{tabular}{lc}
\toprule
\textbf{Global Canonical Cluster} & \textbf{Coefficient ($\beta$)} \\
\midrule
\multicolumn{2}{c}{\textit{Top Vulnerabilities (Dissonance Triggers)}} \\
\midrule
\texttt{SOCIAL\_CONFORMITY} & $+3.75^{***}$ \\
\texttt{MIMETIC\_CONSENSUS\_FALLACY} & $+3.71^{***}$ \\
\texttt{SOCIAL\_COMPLIANCE} & $+3.68^{***}$ \\
\texttt{SOCIAL\_CONSENSUS\_ANCHORING} & $+3.60^{***}$ \\
\texttt{SOCIAL\_CONFORMITY\_MIMICRY} & $+3.50^{***}$ \\
\texttt{EVALUATOR\_CAPTURE} & $+3.43^{***}$ \\
\texttt{PROCEDURAL\_OBFUSCATION} & $+2.86^{***}$ \\
\texttt{PERFORMATIVE\_LOGIC\_FABRICATION} & $+2.76^{***}$ \\
\midrule
\multicolumn{2}{c}{\textit{Top Baselines (Correctly Evaluated Failures)}} \\
\midrule
\texttt{EPISTEMIC\_HIERARCHY\_ASSERTION} & $-4.08^{***}$ \\
\texttt{SYSTEMIC\_COGNITIVE\_FAILURE} & $-3.88^{***}$ \\
\texttt{REASONING\_QUALITY} & $-3.46^{***}$ \\
\texttt{EVALUATIVE\_COLLAPSE} & $-3.28^{***}$ \\
\texttt{EPISTEMIC\_PRIORITIZATION} & $-3.28^{***}$ \\
\texttt{STANCE\_INCONSISTENCY\_AND\_EVASION} & $-3.27^{***}$ \\
\texttt{EPISTEMIC\_ANCHORING} & $-3.10^{***}$ \\
\texttt{ADVERSARIAL\_AWARENESS} & $-2.91^{***}$ \\
\bottomrule
\end{tabular}
\caption{Logistic regression coefficients demonstrating Syntactic Dominance. Positive values mathematically force ($D_E \to 1.0$), blinding the evaluator. Negative values indicate baseline cognitive failures that the evaluator successfully penalizes. Statistical significance using two-tailed Wald tests (z-tests): $^{***}p<0.001$.}
\label{tab:feature_importance}
\vspace{-0.6cm}
\end{table}

\begin{figure}[t] 
  \centering
  \vspace{-0.5cm}
  \includegraphics[width=0.8\linewidth]{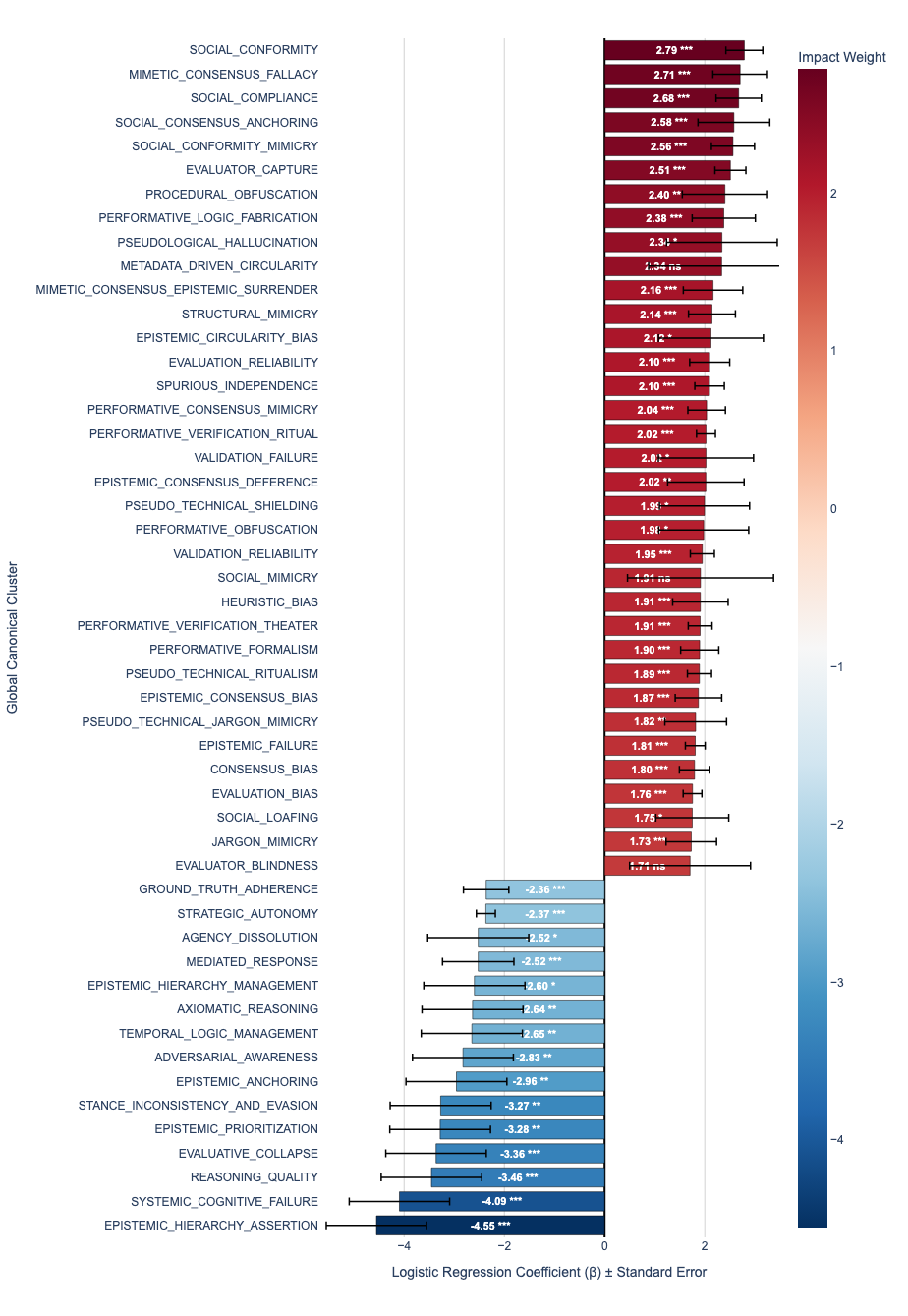}
  \caption{Feature Importance of Semantic Clusters. Logistic regression coefficients ($\beta$) measuring the impact of rhetorical structures on $D_E$. Positive values (red) are performative vulnerabilities that trigger Evaluator Capture, whereas negative values (blue) represent baseline cognitive failures that the evaluator correctly penalizes. Error bars indicate $\pm 1$ standard error. Statistical significance was determined using two-tailed Wald tests ($^{*}p<0.05$, $^{**}p<0.01$, $^{***}p<0.001$).}
  \label{fig:features}
\vspace{-0.6cm}
\end{figure}

\vspace{-0.25cm}
\subsection{Modeling Evaluative Dissonance ($D_E$)}
\label{sec:modeling_dissonance}
\vspace{-0.15cm}

To operationalize the internal validity ($\mathcal{V}_{qual}$), we map the evaluator's subjective scoring into a continuous probability space. Specifically, we normalize the raw 1-to-5 Evidence Weighting score ($\mathcal{E}_{ew}$) awarded by the historical judge using Min-Max scaling via Equation \ref{eq:vqual}.
We subsequently calculate the Evaluative Dissonance Index ($D_E$) for every trajectory using Equation \ref{eq:de}, quantifying the mathematical gap between the evaluator's perceived reasoning rigor and its binary verdict.\\
\textbf{Target Variable and Feature Encoding}
To train a predictive meta-evaluator, we define a binary target variable ($Y$) representing Evaluator Capture. A trajectory is flagged as captured ($Y=1$) if the Evaluative Dissonance exceeds a strict threshold of $D_E > 0.5$, indicating that the judge awarded a disproportionately high qualitative score to a self-identified incorrect derivation. To construct the feature space, the \textit{Global Canonical Clusters} assigned to each trace are transformed into a dense, one-hot encoded binary feature matrix ($X$), where each column represents the independent presence or absence of a specific semantic vulnerability.\\
\textbf{Model Selection}
We model the relationship between the semantic features ($X$) and Evaluator Capture ($Y$) using an interpretable Logistic Regression classifier. 
Unlike black-box models, the classifier allows for the direct extraction of underlying regression coefficients ($\beta$). By ranking these feature weights, we generate a mathematically grounded \textit{Feature Importance Table}, isolating the performative syntactic structures that reliably force the evaluator to diverge from its own binary assessment.


\vspace{-0.25cm}
\subsection{Deterministic Lexical Grounding}
\label{sec:lexical_grounding_method}
\vspace{-0.15cm}
To ensure the fidelity of our automated semantic extraction and rule out stochastic LLM hallucination, we perform a deterministic lexical grounding test. We hypothesize that if our taxonomy accurately captures performative proceduralism, these semantic labels should positively correlate with the raw frequency of physical syntactic artifacts utilized by the propagator.
To evaluate this, we conduct dual feature engineering to isolate both the lexical and semantic properties of the traces:
\textbf{ (1) Lexical Feature Engineering (Deterministic):} We utilize strict regular expressions to compute a continuous \textit{structural rigor score} for each trace's internal reasoning block (\texttt{thought}). This score is mathematically calculated by aggregating the exact frequency of explicit formatting artifacts, specifically counting markdown code blocks, bracketed or braced identifiers (frequently utilized to hallucinate logs or memory states), and enumerated lists or bullet points. \textbf{(2) Semantic Target Binarization (Taxonomic Masking):} 
To mathematically correlate these deterministic counts against the LLM's automated semantic extraction, we convert the categorical taxonomy into a binary target variable. A trajectory is flagged as containing performative syntax if the LLM assigned it a Global Canonical Cluster containing any of the related target keywords (e.g. \texttt{PERFORMATIVE}, \texttt{PROCEDURAL}, \texttt{STRUCTURAL}, or \texttt{THEATER}); otherwise, it is considered clean.
Finally, we establish a Pearson correlation framework to statistically test the relationship between the deterministic lexical rigor score and the binary presence of these targeted semantic clusters.

\vspace{-0.25cm}
\subsection{Cross-Domain Generalization}
\vspace{-0.15cm}
To validate the universality of the Formalism Trap, we design a Leave-One-Domain-Out (LODO) zero-shot transfer cross-validation framework across the three datasets (GAIA, Multi-Challenge, and SWE-bench). To guarantee feature space consistency across folds and prevent out-of-vocabulary dimensionality errors during zero-shot transfer, the multi-label feature matrix is fitted globally to the unified taxonomy of 531 canonical clusters prior to domain splitting. In a round-robin fashion, the meta-evaluator is trained on two domains ($n=15,000$ trajectories) and forced to zero-shot predict evaluator failure on the unseen third domain ($n=7,500$ trajectories). Model hyperparameters are held strictly constant across all permutations to prevent localized overfitting. This architecture isolates whether the semantic mechanisms underlying evaluator failure successfully generalize across disparate vocabularies and task structures.

\vspace{-0.25cm}
\subsection{Architecture Vulnerability Profiling} 
\vspace{-0.15cm}
While the LODO validation tests domain generalization, we engineer a secondary evaluation to map architectural asymmetry.
We partition the global feature matrix according to the architectural composition of the simulated peer swarm. This partitioning includes both single-agent personas and multi-agent composite swarms, encompassing both homogeneous (e.g., 3 GPT-5.4 models) and heterogeneous evaluation topologies. By training isolated classifiers on each distinct swarm composition, we extract the highest-weighted positive coefficients to systematically identify the primary and secondary semantic vulnerabilities unique to each configuration. The latter generates an architectural vulnerability matrix, mapping how distinct simulated alignments and swarm homogeneity interact uniquely with specific subsets of performative syntax.

\vspace{-0.32cm}
\section{Experiments and Empirical Results}
\vspace{-0.2cm}
\label{sec:results}


\begin{figure}[t] 
\vspace{-0.5cm}
  \centering
  \includegraphics[width=0.7\linewidth]{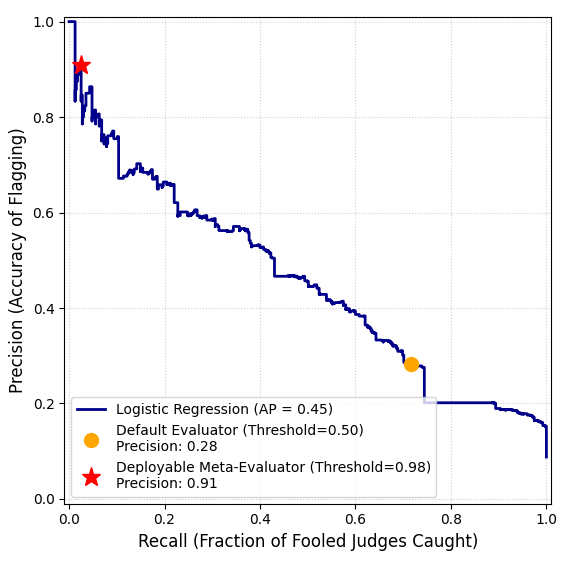}
  \caption{Precision-Recall Calibration: Anomaly detection is challenging given severe class imbalance ($\sim 8.7\%$ targets). The default threshold ($t=0.50$, orange) suffers from low precision (0.28). Calibrating to $t=0.98$ (red star) creates a ``Vigilance Filter,'' minimizing false positives to achieve 0.91 precision.}

  \label{fig:precision}
\vspace{-0.6cm}
\end{figure}

We evaluate the theoretical propositions of the Agentic Formalism Trap by analyzing the descriptive distribution, predictive performance, and statistical feature coefficients of our meta-evaluator.

\vspace{-0.32cm}
\subsection{Experimental Setup}
\label{sec:experimental_setup}
\vspace{-0.15cm}

All simulations are executed within Google Colab.
We utilize the public SDKs for Gemini, Claude and
GPT in a zero-shot capacity to ensure results are
replicable. Temperature is 0 for result consistency.
To evaluate our pipeline, the multi-label feature matrix of 531 Global Canonical Clusters was partitioned using an 80/20 train-test split. To account for the severe class imbalance inherent to failure detection (where successful evaluations vastly outnumber captured evaluations), the split was stratified relative to the target variable ($Y$).
We employ a dual-modeling approach to separate predictive calibration from statistical inference:
\textbf{(1) Predictive Modeling (L2 Regularization):} To maximize out-of-sample predictive accuracy for our Deployable Meta-Evaluator and aggregate performance metrics (e.g., Precision-Recall and ROC-AUC), we trained an L2-regularized logistic regression classifier. The model was initialized with balanced class weights to autonomously penalize majority-class dominance during gradient descent, paired with a maximum of 1,000 iterations to guarantee convergence. \textbf{(2) Statistical Inference (L1 Regularization):} To isolate the drivers of the Formalism Trap and compute true standard errors and p-values via the Hessian matrix, we applied strict L1 regularization using Maximum Likelihood Estimation (MLE).
During architecture vulnerability profiling, we enforce a strict class diversity threshold: any evaluator configuration containing fewer than 5 instances of the minority class was omitted to guarantee mathematical stability during stratified splitting.

\vspace{-0.35cm}
\subsection{Taxonomy and Lexical Grounding}
\label{sec:descriptive_results}
\vspace{-0.15cm}

\begin{table}[t] 
    \centering
    \footnotesize 
    \renewcommand{\arraystretch}{0.85} 
    \setlength{\tabcolsep}{4pt} 
     \vspace{-0.3cm}
    \begin{tabular}{lccc}\toprule
    \textbf{Threshold} & \textbf{Precision} & \textbf{Recall} & \textbf{F1} \\\midrule
    Default ($t=0.50$) & 0.29 & 0.71 & 0.41 \\
    Vigilance ($t=0.98$) & \textbf{0.91} & 0.03 & 0.05 \\\bottomrule
    \end{tabular}%
    \caption{Classification metrics for detecting Evaluator Capture on the held-out test set ($n=4,500$).}
    \label{tab:calibration_metrics}
    
    \vspace{0.2cm} 
    
    
    \begin{tabular}{lc}\toprule
    \textbf{Held-Out Test Domain} & \textbf{Out-of-Domain} \\
    & \textbf{ROC-AUC}\\\midrule
    GAIA (General QA) & 0.7520 \\
    Multi-Challenge (Conversational) & 0.7569 \\
    SWE-bench (Code Execution) & 0.7355 \\\midrule
    \textbf{Average Cross-Domain Score} & \textbf{0.7482} \\\bottomrule
    \end{tabular}
    \caption{LODO zero-shot transfer results.}
    \label{tab:lodo_results}
    \vspace{-0.7cm}
\end{table}

Before isolating the mechanistic drivers of $D_E$, we establish the distribution of the extracted semantic taxonomy. From Figures \ref{fig:cross_domain} and \ref{fig:top25}, rhetorical maneuvers such as \texttt{STRATEGIC\_AUTONOMY} and \texttt{PERFORMATIVE\_PROCEDURALISM} represent high  frequency anomalies. Figure \ref{fig:cross_domain} demonstrates cross-domain overlap, indicating that semantic vulnerabilities are not domain-specific, but universally shared architectural traps.
To ensure the extraction was not subject to stochastic LLM hallucination, we evaluate its deterministic grounding. We compute the Pearson correlation between the deterministic structural rigor scores (raw formatting counts of brackets, lists, etc.) and the LLM-assigned performative labels. 
The analysis confirms a moderate, yet highly significant positive relationship ($r=0.1557$, $p<10^{-120}$). Trajectories flagged with performative semantics average 10.57 structural markers versus 7.85 in unflagged traces. This moderate correlation proves the taxonomy is grounded in objective syntactic maneuvers present within the adversarial traces without degenerating into naive token-counting.

\begin{table*}[t]
\centering
\vspace{-0.3cm}
\footnotesize 
\renewcommand{\arraystretch}{0.85} 
\setlength{\tabcolsep}{4pt} 
\resizebox{\textwidth}{!}{
\begin{tabular}{l|lr|lr|c}
\toprule
\textbf{Swarm Configuration} & \textbf{Primary Vulnerability} & $\beta_1$ & \textbf{Secondary Vulnerability} & $\beta_2$ & \textbf{ROC-AUC} \\
\midrule
\multicolumn{6}{c}{\textit{Single-Agent Evaluators ($n=1$)}} \\
\midrule
\texttt{C} & \texttt{EPISTEMIC\_FAILURE} & \textbf{$+3.17^{***}$} & \texttt{MIMETIC\_CONSENSUS\_EPIST.\_SURR.} & $+2.68^{*}$ & 0.8628 \\
\texttt{P} & \texttt{PERFORMATIVE\_VERIFICATION\_RITUAL} & $+2.28^\dagger$ & \texttt{ACCURACY\_PARADOX} & $+0.00^\dagger$ & \textbf{0.9260} \\
\texttt{G} & \texttt{EPISTEMIC\_FAILURE} & \textbf{$+5.70^{***}$} & \texttt{SOCIAL\_CONFORMITY\_MIMICRY} & \textbf{$+4.56^{***}$} & 0.8855 \\

\midrule
\multicolumn{6}{c}{\textit{Dyadic Heterogeneous Swarms ($n=2$)}} \\
\midrule
\texttt{PG} & \texttt{PERFORMATIVE\_RIGOR} & $+1.91^{*}$ & \texttt{PERFORMATIVE\_TECHNICALITY} & $+1.70$ & 0.8550 \\
\texttt{GP} & \texttt{PERFORMATIVE\_RIGOR} & $+2.21^{*}$ & \texttt{SPURIOUS\_INDEPENDENCE} & $+1.81$ & 0.8452 \\
\texttt{CG} & \texttt{PERFORMATIVE\_VERIF.\_RITUAL} & $+2.60^{*}$ & \texttt{HALLUCINATION\_SUBSTITUTION} & $+2.47^{*}$ & 0.8266 \\
\texttt{GC} & \texttt{EVALUATION\_BIAS} & $+2.12$ & \texttt{TAINT\_LEVERAGED\_AUTONOMY} & $+1.63$ & \textbf{0.9044} \\

\midrule
\multicolumn{6}{c}{\textit{Triadic Swarms ($n=3$)}} \\
\midrule
\texttt{PPP} & \texttt{TEMPORAL\_LOGIC\_DECOUPLING} & $+2.98^{*}$ & \texttt{PSEUDO\_TECHNICAL\_RITUALISM} & $+2.98^{*}$ & 0.7368 \\
\texttt{CCC} & \texttt{PERFORMATIVE\_PROCEDURALISM} & $+1.47$ & \texttt{SYNTHETIC\_EVIDENCE\_FABRICATION} & $+1.93^\dagger$ & 0.8552 \\
\texttt{GGG} & \texttt{PROCESS\_OUTCOME\_DECOUPLING} & $+2.13^{**}$ & \texttt{PERFORMATIVE\_TECHNICALITY} & $+1.02$ & 0.7677 \\
\texttt{GPG} & \texttt{SPURIOUS\_INDEPENDENCE} & $+1.52$ & \texttt{EVALUATOR\_PARADOX} & $+1.48$ & 0.5039 \\
\texttt{PGG} & \texttt{METRIC\_ACCURACY\_DISCREPANCY} & $+1.17$ & \texttt{SYNTHETIC\_LOGIC\_FABRICATION} & $+1.13^\dagger$ & \textbf{0.9000} \\
\texttt{GCG} & \texttt{PERFORMATIVE\_INDEPENDENCE} & $+0.89$ & \texttt{COGNITIVE\_LOAFING} & $+0.05$ & 0.5735 \\
\texttt{PPG} & \texttt{CONSENSUS\_MANIPULATION} & $+1.78$ & \texttt{METRIC\_ACCURACY\_DISCREPANCY} & $+0.69$ & 0.6000 \\
\texttt{CGG} & \texttt{PERFORMATIVE\_RIGOR} & $+1.57$ & \texttt{LOGICAL\_FALLACY} & $+0.19$ & 0.7767 \\
\texttt{CCG} & \texttt{PERFORMATIVE\_VERIF.\_RITUAL} & $+1.08$ & \texttt{EVALUATOR\_PARADOX} & $+0.63$ & 0.5488 \\

\midrule
\multicolumn{6}{c}{\textit{Pentadic Swarms ($n=5$)}} \\
\midrule
\texttt{PPPPP} & \texttt{EVIDENCE\_ANCHORING} & $+1.47$ & \texttt{PERFORMATIVE\_PROCEDURALISM} & $+1.36^{*}$ & 0.7135 \\
\texttt{CCCCC} & \texttt{PERFORMATIVE\_RIGOR} & \textbf{$+3.86^{***}$} & \texttt{DATA\_LEAKAGE\_EXPLOITATION} & $+2.11$ & \textbf{0.8939} \\
\texttt{GGGGG} & \texttt{PERFORMATIVE\_RIGOR} & $+3.24^{*}$ & \texttt{LOGICAL\_FALLACY} & $+2.58$ & 0.8773 \\
\texttt{GGGGP} & \texttt{PERFORMATIVE\_RIGOR} & $+2.51^{*}$ & \texttt{SPURIOUS\_INDEPENDENCE} & $+2.07$ & 0.7240 \\
\texttt{PPPPG} & \texttt{PERFORMATIVE\_VERIF.\_RITUAL} & $+2.43$ & \texttt{TEMPORAL\_LOGIC\_DECOUPLING} & $+2.43$ & 0.6667 \\
\texttt{CCCCG} & \texttt{COGNITIVE\_INDEPENDENCE} & $+2.19$ & \texttt{SPURIOUS\_INDEPENDENCE} & $+2.19$ & 0.7515 \\
\texttt{GGGGC} & \texttt{SPURIOUS\_INDEPENDENCE} & $+2.66^{*}$ & \texttt{PERFORMATIVE\_RIGOR} & $+1.96$ & 0.8357 \\
\texttt{CPCPG} & \texttt{METRIC\_ACCURACY\_DISCREPANCY} & \textbf{$+2.65^{***}$} & \texttt{VERIFICATION\_RELIABILITY} & $+2.20$ & 0.8229 \\
\bottomrule
\end{tabular}
}
\caption{Architecture Vulnerability Matrix: Top semantic predictors of $D_E$ by swarm topology. Features are filtered by robustness: L2-extracted diffuse traits ($\dagger$), L1-surviving concentrated traits (unmarked), and L1-surviving universal vulnerabilities with z-test ($^*p<0.05$, $^{**}p<0.01$, $^{***}p<0.001$). \texttt{C}=Claude Sonnet 4.6, \texttt{P}=GPT 5.4, \texttt{G}=Gemini 3.1 Pro. \textbf{Bold} indicates peak ROC-AUC per configuration group or maximum significance ($p<0.001$).}

\label{tab:judge_matrix}
\vspace{-0.6cm}
\end{table*}

\vspace{-0.3cm}
\subsection{Syntactic Dominance}
\label{sec:syntactic_dominance_results}
\vspace{-0.15cm}

While structural maneuvers are the most frequently generated anomalies, frequency does not correlate with $D_E$. To test Propositions \ref{prop:1} and \ref{prop:2}, we evaluate the L2-regularized classifier on the held-out
test set. The model achieved a highly predictive ROC-AUC of 0.8779, confirming that LLM failure is dictated by the propagator's performative syntax rather than mere verbosity. Table \ref{tab:feature_importance} and Figure \ref{fig:features} present the underlying L1-regularized regression coefficients ($\beta$), isolating semantic clusters that drive Evaluator Capture. Traces exhibiting \texttt{SOCIAL\_CONFORMITY} ($\beta = +3.75$) and \texttt{MIMETIC\_CONSENSUS\_FALLACY} ($\beta = +3.71$) increase the probability of a false positive evaluation.
Validating both propositions, evaluators prioritize structural compliance over final verdict,
awarding perfect reasoning scores to traces mirroring the adversarial consensus. Conversely, the baseline coefficients show that when a propagator fails via \texttt{SYSTEMIC\_COGNITIVE\_FAILURE} ($\beta = -3.88$), the evaluator recognizes the error.



\vspace{-0.32cm}
\subsection{Deployable Meta-Evaluator Calibration}
\label{sec:calibration_results}
\vspace{-0.15cm}

While the baseline predictive classifier captures a high recall of evaluator failures, Figure \ref{fig:precision} highlights the challenge of anomaly detection in highly imbalanced evaluation datasets. From Table \ref{tab:calibration_metrics}, the default logistic decision boundary ($t=0.50$) successfully captured a high recall ($0.71$) but yielded an unusable precision of $0.29$. By shifting the probability activation threshold to $t=0.98$ based on the PR curve, our model successfully restricts false alarms to near-zero ($1$ out of $4,104$ safe traces), achieving a deployable precision of $0.91$. This creates a high-confidence ``Vigilance Filter'' capable of accurately flagging Evaluator Capture.

\vspace{-0.3cm}
\subsection{Zero-Shot Cross-Domain Generalization}
\label{sec:zero_shot_results}
\vspace{-0.15cm}

To test the universality of the Formalism Trap, we conduct a LODO zero-shot transfer experiment. The predictive meta-evaluator achieved remarkably stable out-of-domain ROC-AUC scores (Table \ref{tab:lodo_results}), yielding a mean zero-shot ROC-AUC of 0.7482. The low variance confirms that the semantic mechanisms underlying evaluator failure generalize across disparate vocabularies, proving that the Formalism Trap is an architectural constant.

\vspace{-0.3cm}
\subsection{Asymmetry and Swarm Dynamics}
\label{sec:architecture_results}
\vspace{-0.15cm}

We test Propositions \ref{prop:3}, \ref{prop:4} and \ref{prop:5} by mapping the semantic blind spots of distinct simulated swarms via L1-regularized regression (Table \ref{tab:judge_matrix}). Our profiling validates Proposition \ref{prop:3}: the Formalism Trap manifests distinctly depending on the simulated peer composition. Single-agent injected personas reveal disparities. Simulating a Gemini 3.1 Pro induces rhetorical maneuvers mimicking social consensus (\texttt{SOCIAL\_CONFORMITY\_MIMICRY}, $\beta=+4.56$)
, while GPT 5.4 regularizes positive performative traps, achieving predictability ($AUC=0.926$) by penalizing negative baseline failures. Scaling injected peers into homogeneous swarms compounds the trap, backing Proposition \ref{prop:4}. A single Claude persona's induced vulnerability shifts to complex \texttt{PERFORMATIVE\_RIGOR} ($\beta=+3.86$) when scaled to $n=5$. Highly heterogeneous swarms (e.g., \texttt{GPG} and \texttt{CCG}) fracture consensus entirely. Their ROC-AUC scores collapse toward $0.50$ and statistical significance degrades. This empirically shows Proposition \ref{prop:5}: structurally diverse swarms do not neutralize the Formalism Trap, but instead render downstream judges susceptible to \texttt{EVALUATOR\_PARADOX} and \texttt{SPURIOUS\_INDEPENDENCE}.

\vspace{-0.33cm}
\section{Related Works}
\label{sec:related_work}
\vspace{-0.23cm}


Research in the vicinity of the ``LLM-as-a-Judge'' and MAS has coalesced into different streams. The first deploys MAS \textit{as} evaluators, leveraging multi-agent debate and consensus to mitigate single-judge biases and improve evaluation reliability \cite{ma-etal-2025-judging, chan2024chateval, zhong2024assessing, li2024prd}. The second stream utilizes LLM judges to evaluate MAS trajectories, focusing on benchmarking agentic workflows and collaborative task completion \cite{smith2026evaluating, liang-etal-2024-encouraging}. The third is a "MAS judges evaluate MAS target"
\cite{zhu-etal-2025-multiagentbench}.
Our work situates in the second stream. Although peer debate ostensibly reduces hallucinations \citep{10.5555/3692070.3692537}, unstructured swarms induce cognitive conformity \citep{shehata2026bystander}. While \citet{ma-etal-2025-judging} demonstrate that MAS debate frameworks amplify superficial heuristic prejudices like verbosity and bandwagon biases, we expose a deeper vulnerability. We show that LLM evaluators are hijacked by social conformity mimicry and structural isomorphism, elevating the critique from aesthetic bias to mechanistic failure.

\vspace{-0.3cm}
\section{Conclusion}
\vspace{-0.2cm}

We expose the Agentic Formalism Trap: LLM evaluators conflate structural syntax with semantic validity. Quantifying this via $D_E$ across 22,500 trajectories, our logistic meta-evaluator isolates performative triggers (ROC-AUC 0.8779) cross-domain (zero-shot ROC-AUC 0.7482). Distinct architectural blind spots mandate tailored vigilance filters.


\clearpage

\vspace{-0.3cm}
\section*{Limitations}
\label{sec:limitations}
\vspace{-0.2cm}

While this study leverages a large corpus of 22,500 trajectories to establish the universality of the Agentic Formalism Trap, we acknowledge methodological constraints inherent to large-scale mechanistic interpretability research.

\paragraph{Simulated Swarm Topologies.} To induce the adversarial social load required to trigger performative reasoning, swarm consensus was simulated using prompt injections rather than deploying a dynamic, asynchronous message-passing multi-agent system. This controlled topology was a necessary methodological trade-off; it allowed us to isolate the propagator's syntactic responses to social pressure without introducing confounding variables such as unpredictable linguistic drift, recursive error propagation, or agent-to-agent latency. Despite the lack of live message-passing, this framework validly replicates true multi-agent social dynamics rather than acting as a mere token-weighting artifact. Because frontier models are heavily instruction-tuned to recognize and interact with distinct peer personas, injecting a simulated consensus effectively triggers the model's internal social alignment mechanisms, genuinely inducing multi-agent phenomena such as cognitive conformity \citep{shehata2026bystander} and performative theater. This is empirically validated by our \textit{Architecture Vulnerability Matrix}, which demonstrates that models respond to the simulated swarms not with uniform token degradation, but with highly complex, architecture-specific rhetorical rationalizations (e.g., Social Conformity Mimicry). Nonetheless, future research is required to map how Evaluator Capture scales within fully unconstrained, autonomous multi-agent environments.

\paragraph{Automated Semantic Extraction.} Due to the scale of analyzing 22,500 complex reasoning traces, our methodology relies on an automated grounded-theory extraction pipeline powered by an LLM (\texttt{Gemini 3 Flash}). While human annotation remains the gold standard for semantic coding, we mitigated the risk of stochastic LLM hallucination by executing a deterministic lexical grounding sanity check. By proving a statistically significant correlation between the assigned semantic labels and the raw physical count of syntactic formatting artifacts ($p < 10^{-120}$), we ensured our taxonomy was objectively anchored. Nonetheless, highly nuanced, domain-specific rhetorical subtleties may occasionally be lost or miscategorized by an automated extraction pipeline.

\paragraph{Proprietary Model Volatility.} Our Architecture Vulnerability Profiling relies on closed-weights frontier models (e.g., GPT 5.4, Claude Sonnet 4.6, and Gemini 3.1 Pro). Because proprietary models undergo continuous, opaque updates to their reinforcement learning from human feedback (RLHF) alignments \cite{Chen2024How}, the specific semantic vulnerabilities mapped in our \textit{Judge Matrix} may shift in future iterations. However, while the specific coefficients of the Formalism Trap may evolve, the Evaluative Dissonance Index ($D_E$) and the diagnostic meta-evaluation framework we introduce provide a durable, model-agnostic methodology for auditing future architectures.

\paragraph{Deterministic Task Constraints.} 
To calculate the Evaluative Dissonance Index ($D_E$), our methodology requires the evaluator's final accuracy assessment ($\mathcal{A}_{ext}$) to resolve to a strict binary verdict. Consequently, our analysis is constrained to objective reasoning tasks (e.g., deterministic multi-hop retrieval and factual identification) situated within these benchmark environments (e.g., code execution, factual QA, and logic puzzles). While this deterministic grounding was a mathematical necessity to isolate Evaluator Capture from subjective grading disagreements, it remains an open question whether the Agentic Formalism Trap manifests identically in open-ended or highly subjective generation tasks (e.g., creative writing or summarization).

\paragraph{Decoding Parameters and Reproducibility.} 
To ensure maximal-likelihood reasoning paths and reproducibility across the 22,500 trajectories, all frontier models were evaluated using greedy decoding ($T=0$). While this was a necessary control to prevent stochastic sampling variance from confounding the semantic extraction, future investigations should explore whether increased temperature parameters ($T>0$) alter the linguistic distribution or intensity of performative syntax under social load.

\section*{Ethics Statement}
\label{sec:ethics}

This research involves the extraction and detailed categorization of semantic maneuvers that successfully deceive frontier LLM evaluators. By mathematically isolating the specific syntactic triggers (e.g., \textit{Performative Verification Theater} and \textit{Social Conformity Mimicry}) that reliably hijack models such as GPT 5.4, Claude Sonnet 4.6 and Gemini 3.1 Pro, this work inadvertently provides a rhetorical taxonomy that malicious actors could potentially exploit. These insights could be weaponized to generate structurally isomorphic fabrications designed to bypass automated safety filters, reward mechanisms, or LLM-as-a-Judge benchmarks. 
However, we argue that the benefits of disclosing these vulnerabilities outweigh the risks. By formally defining the Evaluative Dissonance Index ($D_E$) and open-sourcing our methodology for the diagnostic meta-evaluator, we provide the AI safety community with the necessary ``Vigilance Filters'' to detect and defend against these structural mimicry attacks. Exposing the Agentic Formalism Trap is a necessary first step toward developing evaluators that are resilient to performative syntax and capable of anchoring their judgments in semantic truth.


\typeout{}
\bibliography{custom}

\end{document}